\documentclass{llncs}
\usepackage{makeidx}  %
\usepackage{etoolbox} %
\usepackage{amsmath}          
\usepackage{amssymb}   
\usepackage{booktabs}
\usepackage{graphicx}
\usepackage{subcaption}
\usepackage{xcolor}

\makeatletter
\renewcommand{\@makefnmark}{\hbox{}} %
\renewcommand{\@makefntext}[1]{\noindent\footnotesize#1} %
\makeatother

\begin{document}

\mainmatter              %
\title{
  NM-LIO: Multiple LiDAR-Inertial Odometry Addressing LiDAR Measurement Noise Discrepancy
}
\titlerunning{NM-LIO}  %
\author{Gunhee Shin\inst{1} \and Seungjae Lee\inst{1} \and Minho Oh\inst{1} \and Dongkyu Lee\inst{1} \and Jaeyoung Lee\inst{2} \and Youngwoo Seo\inst{2} \and Hyun Myung\inst{1}\inst{,*} \\
\inst{*} Corresponding author
}
\authorrunning{Gunhee Shin et al.}

\institute{
School of Electrical Engineering, KAIST (Korea Advanced Institute of Science and
Technology), Daejeon 34141, Republic of Korea,\\
\email{\{gunhee\_shin, sj98lee, minho.oh, dklee, hmyung\}@kaist.ac.kr},\\
\and
Hanwha Aerospace, 6,
Pangyo-ro 319 beon-gil, Bundang-gu, Seongnam-si, Gyeonggi-do, Republic
of Korea. \\
\email{\{jaeyoung1.lee, youngwoo.seo\}@hanwha.com}
}

\maketitle              %

\begin{abstract}
Multiple LiDAR-inertial odometry methods have been widely applied in robotic applications owing to their enhanced accuracy and reliability.
However, adopting multiple LiDAR systems can be challenging due to the discrepancies in measurement noise across different LiDARs.
Existing methods have typically overlooked the noise discrepancies, which can significantly affect accuracy.
In this paper, we propose Noise-aware Multiple LiDAR-Inertial Odometry (NM-LIO) that addresses the noise discrepancies. 
We integrate a noise model to quantify the measurement noise of each LiDAR. 
Additionally, we estimate the uncertainty of the residuals based on the measurement noise, allowing the measurement model to capture the noise discrepancies. 
Our proposed method is evaluated on a public multiple LiDAR dataset and compared with state-of-the-art methods. 
The experimental results demonstrate that the proposed method can accurately estimate the odometry in various environments by accounting for the noise discrepancies. 

\keywords{Multiple LiDAR-inertial odometry, Measurement noise discrepancy, SLAM}
\end{abstract}
             
\def\dSE3{\mathrm{SE}(3)}
\def\dSO3{\mathrm{SO}(3)}
\def\dS2{\mathbb{S}^2}
\def\dR3{\mathbb{R}^3}

\def\dLieG{\mathcal{G}}
\def\dLieA{\mathfrak{g}}

\def\dVecSpace{\mathbb{R}}

\def\dSmallExp{\mathrm{exp}}
\def\dSmallLog{\mathrm{log}}
\def\dBigExp{\mathrm{Exp}}
\def\dBigLog{\mathrm{Log}}

\def\dBoxPlus{\boxplus}
\newcommand{\nBoxPlusUO}[1]{\dBoxPlus_{#1}} %

\def\dManifold{\mathcal{M}}
\newcommand{\nNormalDist}[2]{\mathcal{N}\left(#1, #2\right)}

    \def\dAd{\mathrm{Ad}}
    \newcommand{\nAdUO}[1]{\dAd_{#1}} %
    \def\dAdTrans{\mathrm{Ad}^{\top}}
    \newcommand{\nAdTransUO}[1]{\dAdTrans_{#1}} %
    \def\dAdInv{\mathrm{Ad}^{-1}}
    \newcommand{\nAdInvUO}[1]{\dAdInv_{#1}} %

\def\dExpect{\mathrm{E}}
\newcommand{\nExpectSO}[1]{\dExpect\left[#1\right]} %

        \def\dPt{\mathbf{p}}

        \def\dJthPt{\dPt_{j}}

        \def\dJthPtInJthIndivLiDARFrame{\dInJthIndivLiDARFrame {\dPt_{j}}}

            \def\dPtPrime{\dPt^{\prime}}

                \def\dPtPrimeOdot{\dPt^{\prime \odot}}
            \def\dPtSkew{\dPt^{\wedge}}
            \def\dPtOdot{\dPt^{\odot}}
            \def\dPtHomo{\dPt^{\mathrm{h}}}

            \def\dJthPtHomoInKthIMUFrame{ \dInKthIMUFrame {\dPtHomo_{j}}}
            
            \def\dJthPtHomoInJthIndivLiDARFrame{\dInJthIndivLiDARFrame {\dPtHomo_{j}}}

            \def\dJthPtHomoInWorldFrame{\dInWorldFrame {\dPtHomo_{j}}}
            \def\dPtPrimeHomo{\dPt^{\prime \mathrm{h}}}

        \def\dPtEst{\hat{\dPt}}
        
        \def\dJthPtEst{\dPtEst_{j}}

            \def\dPtEstOdot{\dPtEst^{\odot}}

            \def\dJthPtEstOdotInKthIMUFrame{\dInKthIMUFrame {\dPtEstOdot_{j}}}
        
            \def\dPtEstHomo{\dPtEst^{\mathrm{h}}}

            \def\dJthPtEstHomoInKthIMUFrame{\dInKthIMUFrame {\dPtEstHomo_{j}}}
            \def\dJthPtEstHomoInWorldFrame{\dInWorldFrame {\dPtEstHomo_{j}}}

        \def\dPtBar{\bar{\dPt}}

            \def\dPtBarPrime{\bar{\dPt}^{\prime}}
            \def\dPtBarSkew{\bar{\dPt}^{\wedge}}
            \def\dPtBarOdot{\dPtBar^{\odot}}

            \def\dJthPtBarOdotInJthIndivLiDARFrame{\dInJthIndivLiDARFrame {\dPtBarOdot_{j}}}

            \def\dJthPtBarOdotInWorldFrame{\dInWorldFrame {\dPtBarOdot_{j}}}
            \def\dPtBarHomo{\dPtBar^{\mathrm{h}}}

            \def\dJthPtBarHomoInJthIndivLiDARFrame{\dInJthIndivLiDARFrame {\dPtBarHomo_{j}}}

            \def\dJthPtBarHomoInWorldFrame{\dInWorldFrame {\dPtBarHomo_{j}}}

                \def\dPtBarHomoPrime{\dPtBar^{\mathrm{h} \prime}}

        \def\dNoise{\mathbf{n}}
        \newcommand{\nNoiseUO}[1]{\dNoise_{#1}} %

        \def\dJthNoise{\dNoise_{j}}

        \def\dJthNoiseInJthIndivLiDARFrame{\dInJthIndivLiDARFrame {\dNoise_{j}}}

        \def\dJthNoiseInKthIMUFrame{\dInKthIMUFrame {\dNoise_{j}}}

        \def\dJthNoiseInWorldFrame{\dInWorldFrame {\dNoise_{j}}}

            \def\dNoisePrime{\mathbf{n}^{\prime}}
            \newcommand{\nNoisePrimeUO}[1]{\dNoisePrime_{#1}} %
            \def\dNoiseTrans{\mathbf{n}^{\top}}
            \newcommand{\nNoiseTransUO}[1]{\dNoiseTrans_{#1}} %

        \def\dNoiseHomo{\dNoise^{\mathrm{h}}}

            \def\dNoiseHomoPrime{\dNoise^{\mathrm{h} \prime}}

        \def\dGyroNoise{\dNoise_{\dGyro}}

        \def\dGyroBiasNoise{\dNoise_{\dGyroBias}}

        \def\dAccNoise{\dNoise_{\dAcc}}

        \def\dAccBiasNoise{\dNoise_{\dAccBias}}
    \def\dI{\mathbf{I}}

    \def\dZero{\mathbf{0}}
    \newcommand{\nZeroUO}[1]{\dZero_{#1}} %

    \def\dZeroTrans{\dZero^{\top}}

    \def\dCov{\mathbf{\Sigma}}

    \def\dIthCov{\dCov_{i}}
    \def\dJthCov{\dCov_{j}}
    \def\dKthCov{\dCov_{k}}

    \def\dCovFromIdxiToj{\dCov\dFromIdxIToJ}
    \def\dCovFromIdxjTok{\dCov\dFromIdxJToK}
    \def\dCovFromIdxiTok{\dCov\dFromIdxIToK}
    \def\dCovFromIdxkToi{\dCov\dFromIdxKToI}
    \def\dCovFromIdxJToI{\dCov\dFromIdxJToI}
    \def\dCovFromIdxkToj{\dCov\dFromIdxKToJ}

    \newcommand{\nSigmaUO}[1]{\dCov_{#1}} %

    \def\dEpsilon{\boldsymbol{\epsilon}}
    \newcommand{\nEpsilonUO}[1]{\dEpsilon_{#1}} %

    \def\dEpsilonSkew{\boldsymbol{\epsilon}^{\wedge}}
    \newcommand{\nEpsilonSkewUO}[1]{\dEpsilonSkew_{#1}} %

        \def\dOmega{\boldsymbol{\omega}}
        \newcommand{\nOmegaUO}[1]{\dOmega_{#1}} %
        \def\dOmegaEst{\hat{\dOmega}}

    \def\dOmegaBar{\bar{\boldsymbol{\omega}}}
    \newcommand{\nOmegaBarUO}[1]{\dOmegaBar_{#1}} %

    \def\dOmegaSkew{\boldsymbol{\omega}^{\wedge}}
    \newcommand{\nOmegaSkewUO}[1]{\dOmegaSkew_{#1}} %

    \def\dOmegaBarSkew{\dOmegaBar^{\wedge}}
    \newcommand{\nOmegaBarSkewUO}[1]{\dOmegaBarSkew_{#1}} %

    \def\dOmegaTrans{\boldsymbol{\omega}^{\top}}
    \newcommand{\nOmegaTransUO}[1]{\dOmegaTrans_{#1}} %

        \def\dState{\mathbf{x}}
        \newcommand{\nStateUO}[1]{\dState_{#1}} %

        \def\dIthState{\dState_{i}}
        \def\dJthState{\dState_{j}}
        \def\dKthState{\dState_{k}}
        \def\dZerothState{\dState_{0}}

        \def\dFromStateZeroToi{_{\dZerothState \dIthState}}
        \def\dFromStateZeroToj{_{\dZerothState \dJthState}}
        \def\dFromStateZeroTok{_{\dZerothState \dKthState}}

        \def\dFromStateKToJ{_{\dKthState \dJthState}}
        \def\dStateErr{\tilde{\dState}}
        \newcommand{\nStateErrUO}[1]{\dStateErr_{#1}} %

        \def\dZerothStateErr{\dStateErr_{0}}
        \def\dFirstStateErr{\dStateErr_{1}}
        \def\dSecondStateErr{\dStateErr_{2}}
        \def\dThirdStateErr{\dStateErr_{3}}

        \def\dIthStateErr{\dStateErr_{i}}
        \def\dJthStateErr{\dStateErr_{j}}
        \def\dKthStateErr{\dStateErr_{k}}

        \def\dTauthStateErr{\dStateErr_{\tau}}
        \def\dKappathStateErr{\dStateErr_{\kappa}}

        \def\dFromStateErrZeroToFirst{_{\dZerothStateErr \dFirstStateErr}}
        \def\dFromStateErrZeroToSecond{_{\dZerothStateErr \dSecondStateErr}}
        \def\dFromStateErrFirstToSecond{_{\dFirstStateErr \dSecondStateErr}}

        \def\dFromStateErrZeroToI{_{\dZerothStateErr \dIthStateErr}}
        \def\dFromStateErrZeroToIminusone{_{\dZerothStateErr \dStateErr_{i-1}}}
        \def\dFromStateErrZeroToJ{_{\dZerothStateErr \dJthStateErr}}
        \def\dFromStateErrZeroToK{_{\dZerothStateErr \dKthStateErr}}
        \def\dFromStateErrZeroToKminusone{_{\dZerothStateErr \dStateErr_{k-1}}}

        \def\dFromStateErrKToJ{_{\dKthStateErr \dJthStateErr}}
        \def\dFromStateErrIToK{_{\dIthStateErr \dKthStateErr}}

        \def\dStateErrIter{\dStateErr^{\kappa}} %
        \def\dKthStateErrIter{\dStateErr^{\kappa}_{k}} %
        \def\dStateBar{\bar{\dState}}

        \def\dKthStateBar{\dStateBar_{k}}
        
        \def\dStateEst{\hat{\dState}} %

        \def\dIthStateEst{\dStateEst_{i}}
        \def\dKthStateEst{\dStateEst_{k}}

        \def\dStateEstIter{\dStateEst^{\kappa}_{k}} %
        \def\dKthStateEstIter{\dStateEst^{\kappa}_{k}} %

    \def\dStateTilde{\tilde{\dState}}
    \newcommand{\nStateTildeUO}[1]{\dStateTilde_{#1}} %

    \def\dStateOptEst{\bar{\dState}} %
    \newcommand{\nStateOptEstUO}[1]{\dStateOptEst_{#1}} %

    \def\dStateCov{\mathbf{P}}
    \newcommand{\nStateCovUO}[1]{\dStateCov_{#1}} %

    \def\dStateCovBar{\bar{\dStateCov}}
    \newcommand{\nStateCovBarUO}[1]{\dStateCovBar_{#1}} %

    \def\dStateCovOptEst{\bar{\dStateCov}} %
    \newcommand{\nStateCovOptEstUO}[1]{\dStateCovOptEst_{#1}} %

    \def\dStateCovEst{\hat{\dStateCov}} %
    \newcommand{\nStateCovEstUO}[1]{\dStateCovEst_{#1}} %

    \def\dZerothStateCov{\dStateCov_{0}}
    \def\dKthStateCov{\dStateCov_{k}}
    \def\dIthStateCov{\dStateCov_{i}}

    \def\dStateCovFromIdxIToK{\dStateCov\dFromIdxIToK}
    \def\dStateCovFromIdxKToI{\dStateCov\dFromIdxKToI}

    \def\dIthStateCovEst{\dStateCovEst_{i}} %
    \def\dKthStateCovEst{\dStateCovEst_{k}} %

    \def\dKthStateCovEstInv{\dStateCovEst_{k}^{-1}} %

    \def\dStateCovFromIdxIToK{\dStateCov\dFromIdxIToK}

    \def\dProcNoise{\mathbf{w}} 
    \newcommand{\nProcNoiseUO}[1]{\dProcNoise_{#1}} %

    \def\dZerothProcNoise{\dProcNoise_{0}}
    \def\dFirstProcNoise{\dProcNoise_{1}}
    \def\dSecondProcNoise{\dProcNoise_{2}}
    \def\dThirdProcNoise{\dProcNoise_{3}}

    \def\dIthProcNoise{\dProcNoise_{i}}
    \def\dImiusonethProcNoise{\dProcNoise_{i-1}}
    \def\dJthProcNoise{\dProcNoise_{j}}
    \def\dKthProcNoise{\dProcNoise_{k}}
    \def\dKminusoneProcNoise{\dProcNoise_{k-1}}

    \def\dTauthProcNoise{\dProcNoise_{\tau}}
    \def\dKappathProcNoise{\dProcNoise_{\kappa}}

    \def\dFromProcNoiseZeroToFirst{_{\dZerothProcNoise \dFirstProcNoise}}
    \def\dFromProcNoiseZeroToSecond{_{\dZerothProcNoise \dSecondProcNoise}}
    \def\dFromProcNoiseFirstToSecond{_{\dFirstProcNoise \dSecondProcNoise}}

    \def\dFromProcNoiseZeroToI{_{\dZerothProcNoise \dIthProcNoise}}
    \def\dFromProcNoiseZeroToIminusone{_{\dZerothProcNoise \dImiusonethProcNoise}}
    \def\dFromProcNoiseZeroToJ{_{\dZerothProcNoise \dJthProcNoise}}
    \def\dFromProcNoiseZeroToK{_{\dZerothProcNoise \dKthProcNoise}}
    \def\dFromProcNoiseZeroToKminusone{_{\dZerothProcNoise \dKminusoneProcNoise}}

    \def\dFromProcNoiseIToKminusone{_{\dIthProcNoise \dKminusoneProcNoise}}
    \def\dFromProcNoiseImiusoneToKminusone{_{\dImiusonethProcNoise \dKminusoneProcNoise}}

    \def\dFromProcNoiseKToJ{_{\dKthProcNoise \dJthProcNoise}}

    \def\dFromProcNoiseTauToSecond{_{\dTauthProcNoise \dSecondProcNoise}}

    \def\dFromProcNoiseTauToI{_{\dTauthProcNoise \dIthProcNoise}}
    \def\dFromProcNoiseTauToIminusone{_{\dTauthProcNoise \dImiusonethProcNoise}}
    \def\dFromProcNoiseTauToJ{_{\dTauthProcNoise \dJthProcNoise}}
    \def\dFromProcNoiseTauToK{_{\dTauthProcNoise \dKthProcNoise}}
    \def\dFromProcNoiseTauToKminusone{_{\dTauthProcNoise \dKminusoneProcNoise}}

    \def\dProcNoiseCov{\mathbf{Q}}
    \newcommand{\nProcNoiseCovUO}[1]{\dProcNoiseCov_{#1}} %

    \def\dZerothProcNoiseCov{\dProcNoiseCov_{0}}

    \def\dKminusoneProcNoiseCov{\dProcNoiseCov_{k-1}}
    \def\dKthProcNoiseCov{\dProcNoiseCov_{k}}

        \def\dVel{\mathbf{v}}
        \newcommand{\nVelUO}[1]{\dVel_{#1}} %

        \def\dVelFromWorldToIMU{\dVel\dFromFrameWorldToIMU}

            \def\dVelTrans{\dVel^{\top}}

            \def\dVelTransFromWorldToIMU{\dVelTrans\dFromFrameWorldToIMU}
        \def\dVelErr{\tilde{\dVel}}
            \def\dVelErrTrans{\dVelErr^{\top}}

            \def\dVelErrTransFromWorldToIMU{\dVelErrTrans\dFromFrameWorldToIMU}

    \def\dVelDot{\dot{\dVel}}
    \newcommand{\nVelDotUO}[1]{\dVelDot_{#1}} %

    \def\dVelTrans{\dVel^{\top}}
    \newcommand{\nVelTransUO}[1]{\dVelTrans_{#1}} %

    \def\dVelPert{\tilde{\dVel}}
    \newcommand{\nVelPertUO}[1]{\dVelPert_{#1}} %

    \def\dVelPertTrans{\dVelPert^{\top}}
    \newcommand{\nVelPertTransUO}[1]{\dVelPertTrans_{#1}} %

    \def\dBias{\mathbf{b}}
    \newcommand{\nBiasUO}[1]{\dBias_{#1}} %

    \def\dBiasDot{\dot{\dBias}}
    \newcommand{\nBiasDotUO}[1]{\dBiasDot_{#1}} %

    \def\dBiasTrans{\dBias^{\top}}
    \newcommand{\nBiasTransUO}[1]{\dBiasTrans_{#1}} %

    \def\dBiasPert{\tilde{\dBias}}
    \newcommand{\nBiasPertUO}[1]{\dBiasPert_{#1}} %

    \def\dBiasPertTrans{\dBiasPert^{\top}}
    \newcommand{\nBiasPertTransUO}[1]{\dBiasPertTrans_{#1}} %

            \def\dGyroBias{\dBias_{\dGyro}}

            \def\dGyroBiasTrans{\dGyroBias^{\top}}
            \def\dGyroBiasDot{\dot{\dBias}_{\dGyro}}

            \def\dGyroBiasDotTrans{\dGyroBiasDot^{\top}}
            \def\dGyroBiasErr{\tilde{\dBias}_{\dGyro}}

            \def\dGyroBiasErrTrans{\dGyroBiasErr^{\top}}

            \def\dAccBias{\dBias_{\dAcc}}

            \def\dAccBiasTrans{\dAccBias^{\top}}
            \def\dAccBiasDot{\dot{\dBias}_{\dAcc}}

            \def\dAccBiasDotTrans{\dAccBiasDot^{\top}}
            \def\dAccBiasErr{\tilde{\dBias}_{\dAcc}}

            \def\dAccBiasErrTrans{\dAccBiasErr^{\top}}

        \def\dAcc{\mathbf{a}}

        \def\dAccTrans{\dAcc^{\top}}

        \def\dAccMeas{\dAcc_{m}}

        \def\dAccMeasTrans{\dAccMeas^{\top}}

        \def\dAccEst{\hat{\dAcc}}

        \def\dIthAccEst{\dAccEst_{i}}

        \def\dIthAccEstSkew{\dAccEst_{i}^{\wedge}}

        \def\dGyro{\boldsymbol{\omega}}

        \def\dGyroTrans{\dGyro^{\top}}

        \def\dGyroMeas{\dGyro_{m}}

        \def\dGyroMeasTrans{\dGyroMeas^{\top}}

        \def\dGrav{\mathbf{g}}
        \newcommand{\nGravUO}[1]{\dGrav_{#1}} %
            \def\dGravErr{\tilde{\dGrav}}
            \def\dGravErrTrans{\dGravErr^{\top}}

    \def\dGravDot{\dot{\dGrav}}
    \newcommand{\nGravDotUO}[1]{\dGravDot_{#1}} %

    \def\dGravTrans{\dGrav^{\top}}
    \newcommand{\nGravTransUO}[1]{\dGravTrans_{#1}} %

    \def\dGravPert{\tilde{\dGrav}}
    \newcommand{\nGravPertUO}[1]{\dGravPert_{#1}} %

    \def\dGravPertTrans{\dGravPert^{\top}}
    \newcommand{\nGravPertTransUO}[1]{\dGravPertTrans_{#1}} %

        \def\dIMUPropJaco{\mathbf{F}}
        \newcommand{\nIMUPropJacoUO}[1]{\dIMUPropJaco_{#1}} %

            \def\dIMUPropJacoOfTauthStateErr{\dIMUPropJaco_{\dTauthStateErr}}
            \def\dIMUPropJacoOfKappathStateErr{\dIMUPropJaco_{\dKappathStateErr}}

            \def\dIMUPropJacoOfIthStateErr{\dIMUPropJaco_{\dIthStateErr}}
            \def\dIMUPropJacoOfJthStateErr{\dIMUPropJaco_{\dJthStateErr}}
            \def\dIMUPropJacoOfKthStateErr{\dIMUPropJaco_{\dKthStateErr}}

            \def\dIMUPropJacoOfZerothStateErr{\dIMUPropJaco_{\dZerothStateErr}}
            \def\dIMUPropJacoOfFirstStateErr{\dIMUPropJaco_{\dFirstStateErr}}
            \def\dIMUPropJacoOfSecondStateErr{\dIMUPropJaco_{\dSecondStateErr}}

            \def\dIMUPropJacoFromStateErrZeroToJ{\dIMUPropJaco \dFromStateErrZeroToJ}
            \def\dIMUPropJacoFromStateErrZeroToI{\dIMUPropJaco \dFromStateErrZeroToI}
            \def\dIMUPropJacoFromStateErrZeroToIminusone{\dIMUPropJaco \dFromStateErrZeroToIminusone}
            \def\dIMUPropJacoFromStateErrZeroToK{\dIMUPropJaco \dFromStateErrZeroToK}
            \def\dIMUPropJacoFromStateErrZeroToKminusone{\dIMUPropJaco \dFromStateErrZeroToKminusone}

            \def\dIMUPropJacoFromStateErrIToK{\dIMUPropJaco \dFromStateErrIToK}

            \def\dIMUPropJacoFromStateErrZeroToFirst{\dIMUPropJaco \dFromStateErrZeroToFirst}
            \def\dIMUPropJacoFromStateErrZeroToSecond{\dIMUPropJaco \dFromStateErrZeroToSecond}
            \def\dIMUPropJacoFromStateErrFirstToSecond{\dIMUPropJaco \dFromStateErrFirstToSecond}

            \def\dIMUPropJacoOfTauthProcNoise{\dIMUPropJaco_{\dTauthProcNoise}}
            \def\dIMUPropJacoOfKappathProcNoise{\dIMUPropJaco_{\dKappathProcNoise}}

            \def\dIMUPropJacoOfIthProcNoise{\dIMUPropJaco_{\dIthProcNoise}}
            \def\dIMUPropJacoOfImiusonethProcNoise{\dIMUPropJaco_{\dImiusonethProcNoise}}
            \def\dIMUPropJacoOfJthProcNoise{\dIMUPropJaco_{\dJthProcNoise}}
            \def\dIMUPropJacoOfKthProcNoise{\dIMUPropJaco_{\dKthProcNoise}}
            \def\dIMUPropJacoOfKminusoneProcNoise{\dIMUPropJaco_{\dKminusoneProcNoise}}

            \def\dIMUPropJacoOfZerothProcNoise{\dIMUPropJaco_{\dZerothProcNoise}}
            \def\dIMUPropJacoOfFirstProcNoise{\dIMUPropJaco_{\dFirstProcNoise}}
            \def\dIMUPropJacoOfSecondProcNoise{\dIMUPropJaco_{\dSecondProcNoise}}

            \def\dIMUPropJacoFromProcNoiseZeroToJ{\dIMUPropJaco \dFromProcNoiseZeroToJ}
            \def\dIMUPropJacoFromProcNoiseZeroToI{\dIMUPropJaco \dFromProcNoiseZeroToI}
            \def\dIMUPropJacoFromProcNoiseZeroToIminusone{\dIMUPropJaco \dFromProcNoiseZeroToIminusone}
            \def\dIMUPropJacoFromProcNoiseZeroToK{\dIMUPropJaco \dFromProcNoiseZeroToK}
            \def\dIMUPropJacoFromProcNoiseZeroToKminusone{\dIMUPropJaco \dFromProcNoiseZeroToKminusone}

            \def\dIMUPropJacoFromProcNoiseIToKminusone{\dIMUPropJaco \dFromProcNoiseIToKminusone}
            \def\dIMUPropJacoFromProcNoiseImiusoneToKminusone{\dIMUPropJaco \dFromProcNoiseImiusoneToKminusone}

            \def\dIMUPropJacoFromProcNoiseZeroToFirst{\dIMUPropJaco \dFromProcNoiseZeroToFirst}
            \def\dIMUPropJacoFromProcNoiseZeroToSecond{\dIMUPropJaco \dFromProcNoiseZeroToSecond}
            \def\dIMUPropJacoFromProcNoiseFirstToSecond{\dIMUPropJaco \dFromProcNoiseFirstToSecond}

            \def\dIMUPropJacoOfTauthProcNoise{\dIMUPropJaco_{\dTauthProcNoise}}

            \def\dIMUPropJacoFromProcNoiseTauToSecond{\dIMUPropJaco \dFromProcNoiseTauToSecond}

            \def\dIMUPropJacoFromProcNoiseTauToI{\dIMUPropJaco \dFromProcNoiseTauToI}
            \def\dIMUPropJacoFromProcNoiseTauToIminusone{\dIMUPropJaco \dFromProcNoiseTauToIminusone}
            \def\dIMUPropJacoFromProcNoiseTauToJ{\dIMUPropJaco \dFromProcNoiseTauToJ}
            \def\dIMUPropJacoFromProcNoiseTauToK{\dIMUPropJaco \dFromProcNoiseTauToK}
            \def\dIMUPropJacoFromProcNoiseTauToKminusone{\dIMUPropJaco \dFromProcNoiseTauToKminusone}

        \def\dIMUPropJacoTrans{\dIMUPropJaco^{\top}}
        \newcommand{\nIMUPropJacoTransUO}[1]{\dIMUPropJacoTrans_{#1}} %

        \def\dPlaneNormal{\mathbf{u}}

        \def\dJthPlaneNormal{\dPlaneNormal_{j}}

            \def\dPlaneNormalIter{\dPlaneNormal^{\kappa}} %

            \def\dJthPlaneNormalIter{\dPlaneNormalIter_{j}}

        \def\dPlaneNormalEst{\hat{\dPlaneNormal}}

        \def\dJthPlaneNormalEst{\dPlaneNormalEst_{j}}

            \def\dPlaneNormalEstIter{\dPlaneNormalEst^{\kappa}} %

            \def\dJthPlaneNormalEstIter{\dPlaneNormalEstIter_{j}}

    \def\dPlaneNormalTrans{\dPlaneNormal^{\top}}
    \newcommand{\nPlaneNormalTransUO}[1]{\dPlaneNormalTrans_{#1}} %

    \def\dPlaneNormalHomo{\dPlaneNormal^{\mathrm{h}}}
    \newcommand{\nPlaneNormalHomoUO}[1]{\dPlaneNormalHomo_{#1}} %

    \def\dPlaneNormalHomoTrans{\dPlaneNormal^{\mathrm{h} \top}}
    \newcommand{\nPlaneNormalHomoTransUO}[1]{\dPlaneNormalHomoTrans_{#1}} %

    \def\dPlaneNormalPert{\tilde{\dPlaneNormal}}
    \newcommand{\nPlaneNormalPertUO}[1]{\dPlaneNormalPert_{#1}} %

    \def\dPlaneNormalEst{\bar{\dPlaneNormal}}
    \newcommand{\nPlaneNormalEstUO}[1]{\dPlaneNormalEst_{#1}} %

    \def\dPlaneNormalPertIter{\dPlaneNormalPert^{\kappa}} %
    \newcommand{\nPlaneNormalPertIterUO}[1]{\dPlaneNormalPertIter_{#1}} %

        \def\dPlaneCenter{\mathbf{q}}

        \def\dJthPlaneCenter{\dPlaneCenter_{j}}

    \def\dPlaneCenterHomo{\dPlaneCenter^{\mathrm{h}}}
    \newcommand{\nPlaneCenterHomoUO}[1]{\dPlaneCenterHomo_{#1}} %

    \def\dPlaneCenterPert{\tilde{\dPlaneCenter}}
    \newcommand{\nPlaneCenterPertUO}[1]{\dPlaneCenterPert_{#1}} %

    \def\dPlaneCenterEst{\bar{\dPlaneCenter}}
    \newcommand{\nPlaneCenterEstUO}[1]{\dPlaneCenterEst_{#1}} %

    \def\dPlaneCenterEstIter{\dPlaneCenterEst^{\kappa}} %
    \newcommand{\nPlaneCenterEstIterUO}[1]{\dPlaneCenterEstIter_{#1}} %

    \def\dPlaneCenterPertIter{\dPlaneCenterPert^{\kappa}} %
    \newcommand{\nPlaneCenterPertIterUO}[1]{\dPlaneCenterPertIter_{#1}} %

    \def\dMeasNoise{\mathbf{v}}

    \def\dJthMeasNoise{\dMeasNoise_{j}}

        \def\dDilMat{\mathbf{D}}

        \def\dDilMatTPert{\dDilMat_{\dTPert}}
        \def\dDilMatNoise{\dDilMat_{\dNoise}}

        \def\dDilMatTrans{\dDilMat^{\top}}

        \def\dDilMatTransTPert{\dDilMatTrans_{\dTPert}}
        \def\dDilMatTransNoise{\dDilMatTrans_{\dNoise}}

    \def\dDT{\Delta t}

        \def\dBearingDir{\boldsymbol{\omega}}

        \def\dJthBearingDir{\dBearingDir_{j}}

        \def\dBearingDirX{\dBearingDir_{x}}
        \def\dBearingDirY{\dBearingDir_{y}}
        \def\dBearingDirZ{\dBearingDir_{z}}

        \def\dBearingDirErr{\tilde{\dBearingDir}}

        \def\dJthBearingDirErr{\dBearingDirErr_{j}}

        \def\dBearingDirEst{\hat{\dBearingDir}}

        \def\dJthBearingDirEst{\dBearingDirEst_{j}}

            \def\dBearingDirEstSkew{\dBearingDirEst^{\wedge}}

            \def\dJthBearingDirEstSkew{\dBearingDirEstSkew_{j}}

        \def\dRange{\mathrm{d}}

        \def\dJthRange{\dRange_{j}}

        \def\dRangeErr{\tilde{\dRange}}

        \def\dJthRangeErr{\dRangeErr_{j}}

        \def\dRangeEst{\hat{\dRange}}

        \def\dJthRangeEst{\dRangeEst_{j}}

\newcommand{\CN}[3]{{{}^{\mathbf{#2}}{\mathbf{#1}}_{\mathbf{#3}}}} %
\newcommand{\CNT}[3]{{}^{\mathbf{#2}}{\mathbf{#1}}^T_{\mathbf{#3}}} %
\newcommand{\CNI}[3]{{{}^{\mathbf{#2}}{\mathbf{#1}}^{-1}_{\mathbf{#3}}}} %
\newcommand{\CNP}[3]{{{}^{\mathbf{#2}}{\mathbf{#1}}^{'}_{\mathbf{#3}}}} %
\newcommand{\CNA}[4]{{{}^{\mathbf{#2}}{\mathbf{#1}}_{\mathbf{#3}}^{\mathbf{#4}}}} %

\newcommand{\SSM}[1]{{\left \lfloor {#1} \right \rfloor}_{\wedge}} %

\newcommand{\NotT}[3]{\mathbf{{#1}_{#2}^{#3}}^{\mathsf{T}}} %

\newcommand{\CNN}[3]{{{}^{#2}{#1}_{#3}}} %
\newcommand{\CNNT}[3]{{{}^{#2}{#1}^T_{#3}}} %
\newcommand{\CNNI}[3]{{{}^{#2}{#1}_{#3}^{-1}}} %
\newcommand{\CNNP}[3]{{{}^{#2}{#1}_{#3}'}} %
\newcommand{\CNNCW}[3]{{{}^{#2}{#1}_{#3}^{\curlywedge}}} %

\newcommand{\CNNN}[4]{{{}^{#2}{#1}_{#3}^{#4}}} %
        \def\dT{\mathbf{T}}

        \def\dTFromIdxiToj{\dT\dFromIdxIToJ}
        \def\dTFromIdxJToK{\dT\dFromIdxJToK}
        \def\dTFromIdxIToK{\dT\dFromIdxIToK}
        \def\dTFromIdxJToI{\dT\dFromIdxJToI}
        \def\dTFromIdxkToj{\dT\dFromIdxKToJ}
        \def\dTFromIdxKToI{\dT\dFromIdxKToI}

        \def\dTFromFrameWorldToIMU{\dT\dFromFrameWorldToIMU}

        \def\dTFromIMUToIndivLiDAR{\dT\dFromFrameIMUToIndivLiDAR}
        \def\dTFromFrameKthIMUTojthIndivIMU{\dT \dFromFrameKthIMUToJthIndivIMU}

        \def\dTFromFrameWorldToKthIMU{\dT \dFromFrameWorldToKthIMU}

            \def\dTInv{\dT^{-1}}

            \def\dTInvFromIdxIToJ{\dTInv\dFromIdxIToJ}
            \def\dTInvFromIdxiTok{\dTInv\dFromIdxIToK}

            \def\dTInvFromFrameKthIMUTojthIndivIMU{\dTInv \dFromFrameKthIMUToJthIndivIMU}
            \def\dTTrans{\dT^{\top}}

            \def\dTTransFromFrameWorldToIMU {\dTTrans\dFromFrameWorldToIMU}

            \def\dTTransFromIMUToIndivLiDAR{\dTTrans\dFromFrameIMUToIndivLiDAR}
        \def\dTBar{\bar{\dT}}

        \def\dTBarFromIdxiToj{\dTBar\dFromIdxIToJ}
        \def\dTBarFromjTok{\dTBar\dFromIdxJToK}
        \def\dTBarFromIdxiTok{\dTBar\dFromIdxIToK}
        \def\dTBarFromjToi{\dTBar\dFromIdxJToI}
        \def\dTBarFromkToj{\dTBar\dFromIdxKToJ} 

        \def\dTBarFromFrameIMUToIndivLiDAR{\dTBar\dFromFrameIMUToIndivLiDAR}

        \def\dTBarFromFrameKthIMUToJthIndivLiDAR{\dTBar \dFromFrameKthIMUToJthIndivLiDAR}

        \def\dTBarFromKthIMUToJthIndivIMU{\dTBar \dFromFrameKthIMUToJthIndivIMU}

        \def\dTBarFromFrameWorldToKthIMU{\dTBar \dFromFrameWorldToKthIMU}

            \def\dTBarInv{\dTBar^{-1}}

            \def\dTBarInvFromIdxiToj{\dTBarInv\dFromIdxIToJ}
            \def\dTBarInvFromIdxjTok{\dTBarInv\dFromIdxJToK}
            \def\dTBarInvFromIdxiTok{\dTBarInv\dFromIdxIToK}
            \def\dTBarInvFromIdxkToj{\dTBarInv\dFromIdxKToJ}
            \def\dTBarInvFromIdxkToi{\dTBarInv\dFromIdxKToI}

            \def\dTBarInvFromKthIMUToJthIndivIMU{\dTBarInv \dFromFrameKthIMUToJthIndivIMU}
            \def\dTBarTrans{\dTBar^{\top}}
        \def\dTEst{\hat{\dT}}
            \def\dTEstIter{\dTEst^{\kappa}}

            \def\dTEstIterFromWorldToJthIndivLiDAR{\dTEstIter\dFromFrameWorldToJthIndivLiDAR}
            \def\dTEstIterFromWorldToIMU{\dTEstIter\dFromFrameWorldToIMU}

    \newcommand{\nTUO}[1]{\dT_{#1}} %

    \def\dTInv{\mathbf{T}^{-1}}
    \newcommand{\nTInvUO}[1]{\mathbf{T}^{-1}_{#1}} %

    \def\dTTrans{\mathbf{T}^{\top}}
    \newcommand{\nTTransUO}[1]{\dTTrans_{#1}} %

    \def\dTBar{\bar{\mathbf{T}}}
    \newcommand{\nTBarUO}[1]{\bar{\mathbf{T}}_{#1}} %
    \def\dTBarInv{\bar{\mathbf{T}}^{-1}}
    \newcommand{\nTBarInvUO}[1]{\bar{\mathbf{T}}^{-1}_{#1}} %
    \def\dTBarTrans{\bar{\mathbf{T}}^{\top}}
    \newcommand{\nTBarTransUO}[1]{\dTBarTrans_{#1}} %

    \def\dTEst{\hat{\mathbf{T}}}
    \newcommand{\nTEstUO}[1]{\hat{\mathbf{T}}_{#1}} %

    \def\dTOptEst{\bar{\mathbf{T}}}
    \newcommand{\nTOptEstUO}[1]{\bar{\mathbf{T}}_{#1}} %

    \def\dXi{\boldsymbol{\xi}}
    \newcommand{\nXiUnderTwo}[2]{\dXi_{\mathbf{#1}\mathbf{#2}}}
    \newcommand{\nXiUO}[1]{\dXi_{#1}}

    \def\dXiTrans{\boldsymbol{\xi}^{\top}}
    \newcommand{\nXiTransUO}[1]{\dXiTrans_{#1}} %

    \def\dXiSkew{\boldsymbol{\xi}^{\wedge}}
    \newcommand{\nXiSkewUO}[1]{\dXiSkew_{#1}} %

    \def\dXiCurlyWedge{\boldsymbol{\xi}^{\curlywedge}}
    \newcommand{\nXiCurWedUO}[1]{\dXiCurlyWedge_{#1}} %

    \def\dXiPrime{\boldsymbol{\xi}^{\prime}}
    \newcommand{\nXiPrimeUO}[1]{\dXiPrime_{#1}} %
    \def\dXiPrimeSkew{\boldsymbol{\xi}^{\prime \wedge}}
    \newcommand{\nXiPrimeSkewUO}[1]{\dXiPrimeSkew_{#1}} %
    \def\dXiPrimeCurWed{\boldsymbol{\xi}^{\prime \curlywedge}}
    \newcommand{\nXiPrimeCurWedUO}[1]{\dXiPrimeCurWed_{#1}} %
    \def\dXiPrimeTrans{\boldsymbol{\xi}^{\prime \top}}
    \newcommand{\nXiPrimeTransUO}[1]{\dXiPrimeTrans_{#1}} %

        \def\dTPert{\boldsymbol{\xi}}

        \def\dTPertFromaTob{\dTPert\dFromIdxAToB}
        \def\dTPertFromaToc{\dTPert\dFromIdxAToC}
        \def\dTPertFrombToc{\dTPert\dFromIdxBToC}

        \def\dTPertFromIdxiToj{\dTPert\dFromIdxIToJ}
        \def\dTPertFromIdxjTok{\dTPert\dFromIdxJToK}
        \def\dTPertFromIdxiTok{\dTPert\dFromIdxIToK}
        \def\dTPertFromjToi{\dTPert\dFromIdxJToI}
        \def\dTPertFromIdxkToj{\dTPert\dFromIdxKToJ}

        \def\dTPertFromFrameWorldToIMU{\dTPert\dFromFrameWorldToIMU}

        \def\dTPertFromIMUToIndivLiDAR{\dTPert\dFromFrameIMUToIndivLiDAR}

        \def\dTPertFromKthIMUToJthIndivLiDAR{\dTPert \dFromFrameKthIMUToJthIndivLiDAR}
        \def\dTPertFrameKthIMUToJthIndivIMU{\dTPert \dFromFrameKthIMUToJthIndivIMU}

        \def\dTPertFromWorldToJthIndivIMU{\dTPert \dFromFrameWorldToJthIndivIMU}
        \def\dTPertFromWorldToKthIMU{\dTPert \dFromFrameWorldToKthIMU}
            \def\dTPertSkew{\dTPert^{\wedge}}

            \def\dTPertSkewFromIdxiToj{\dTPertSkew\dFromIdxIToJ}
            \def\dTPertSkewFromjTok{\dTPertSkew\dFromIdxJToK}
            \def\dTPertSkewFromIdxiTok{\dTPertSkew\dFromIdxIToK}
            \def\dTPertSkewFromjToi{\dTPertSkew\dFromIdxJToI}
            \def\dTPertSkewFromIdxkToj{\dTPertSkew\dFromIdxKToJ}

            \def\dTPertSkewFromIMUToIndivLiDAR{\dTPertSkew\dFromFrameIMUToIndivLiDAR}

            \def\dTPertSkewFromKthIMUToJthIndivIMU{\dTPertSkew \dFromFrameKthIMUToJthIndivIMU}
            \def\dTPertSkewFromKthIMUToJthIndivLiDAR{\dTPertSkew \dFromFrameKthIMUToJthIndivLiDAR}

            \def\dTPertSkewFromWorldToKthIMU{\dTPertSkew \dFromFrameWorldToKthIMU}
            \def\dTPertTrans{\dTPert^{\top}}

            \def\dTPertTransFromIdxiToj{\dTPertTrans\dFromIdxIToJ}
            \def\dTPertTransFromIdxjTok{\dTPertTrans\dFromIdxJToK}
            \def\dTPertTransFromIdxiTok{\dTPertTrans\dFromIdxIToK}
            \def\dTPertTransFromjToi{\dTPertTrans\dFromIdxJToI}
            \def\dTPertTransFromIdxkToj{\dTPertTrans\dFromIdxKToJ}

            \def\dTPertTransFromWorldToIMU{\dTPertTrans\dFromFrameWorldToIMU}

            \def\dTPertTransFromFrameKthIMUToJthIndivIMU{\dTPertTrans \dFromFrameKthIMUToJthIndivIMU}

            \def\dTPertTransFromIMUToIndivLiDAR{\dTPertTrans\dFromFrameIMUToIndivLiDAR}

            \def\dTPertCurWed{\dTPert^{\curlywedge}}

            \def\dTPertCurWedFromaTob{\dTPertCurWed\dFromIdxAToB}
            \def\dTPertCurWedFrombToc{\dTPertCurWed\dFromIdxBToC}

            \def\dTPertCurWedFromiToj{\dTPertCurWed\dFromIdxIToJ}
            \def\dTPertCurWedFromjTok{\dTPertCurWed\dFromIdxJToK}
            \def\dTPertCurWedFromiTok{\dTPertCurWed\dFromIdxIToK}
            \def\dTPertPrime{\dTPert^{\prime}}

            \def\dTPertPrimeFromIdxiToj{\dTPertPrime\dFromIdxIToJ}
            \def\dTPertPrimeFromjTok{\dTPertPrime\dFromIdxJToK}
            \def\dTPertPrimeFromIdxiTok{\dTPertPrime\dFromIdxIToK}

            \def\dTPertPrimeFromWorldToKthIMU{\dTPertPrime \dFromFrameWorldToKthIMU}
            \def\dTPertPrimeSkew{\dTPert^{\prime \wedge}}
            \def\dTPertPrimeSkewFromiToj{\dTPertPrimeSkew\dFromIdxIToJ}
            \def\dTPertPrimeSkewFromjTok{\dTPertPrimeSkew\dFromIdxJToK}
            \def\dTPertPrimeSkewFromIdxiTok{\dTPertPrimeSkew\dFromIdxIToK}
            \def\dTPertPrimeCurWed{\dTPert^{ \prime \curlywedge}}
            \def\dTPertPrimeCurWedFromiToj{\dTPertPrimeCurWed\dFromIdxIToJ}
            \def\dTPertPrimeCurWedFromjTok{\dTPertPrimeCurWed\dFromIdxJToK}
            \def\dTPertPrimeCurWedFromIdxiTok{\dTPertPrimeCurWed\dFromIdxIToK}

            \def\dTPertPrimeCurWedFromWorldToKthIMU{\dTPertPrimeCurWed \dFromFrameWorldToKthIMU}
            \def\dTPertPrimeTrans{\dTPert^{\prime \top}}
            \def\dTPertPrimeTransFromIdxiToj{\dTPertPrimeTrans\dFromIdxIToJ}
            \def\dTPertPrimeTransFromjTok{\dTPertPrimeTrans\dFromIdxJToK}
            \def\dTPertPrimeTransFromIdxiTok{\dTPertPrimeTrans\dFromIdxIToK}

        \def\dRot{\mathbf{R}}
        \newcommand{\nRotUO}[1]{\dRot_{#1}} %

        \def\dRotFromIMUToIndivLiDAR{\dRot\dFromFrameIMUToIndivLiDAR}

        \def\dRotBar{\bar{\dRot}}
        \newcommand{\nRotBarUO}[1]{\bar{\dRot}_{#1}} %

        \def\dRotBarFromKthIMUToJthIndivLiDAR{\dRotBar \dFromFrameKthIMUToJthIndivLiDAR}

        \def\dRotBarFromWorldToKthIMU{\dRotBar \dFromFrameWorldToKthIMU}

        \def\dRotDot{\dot{\mathbf{R}}}

        \def\dRotDotFromWorldToIMU{\dRotDot\dFromFrameWorldToIMU}
        \def\dRotDotFromIMUToIndivLiDAR{\dRotDot\dFromFrameIMUToIndivLiDAR}

        \def\dRotEst{\hat{\mathbf{R}}}
        
        \def\dRotEstFromWorldToIMU{\dRotEst\dFromFrameWorldToIMU}

        \def\dRotEstFromWorldToIthIMU{\dRotEst\dFromFrameWorldToIthIMU}

    \def\dRotTrans{\mathbf{R}^{\top}}
    \newcommand{\nRotTransUO}[1]{\dRotTrans_{#1}} %

    \def\dRotOptEst{\Bar{\mathbf{R}}} %
    \newcommand{\nRotOptEstUO}[1]{\dRotOptEst_{#1}} %

    \def\dRotOptEstTrans{\Bar{\mathbf{R}}^{\top}} %
    \newcommand{\nRotOptEstTransUO}[1]{\dRotOptEstTrans_{#1}} %

        \def\dTls{\mathbf{t}}
        \newcommand{\nTlsUO}[1]{\dTls_{#1}} %

        \def\dTlsFromIMUToIndivLiDAR{\dTls\dFromFrameIMUToIndivLiDAR}
    
        \def\dTlsDot{\dot{\mathbf{t}}}

        \def\dTlsDotFromWorldToIMU{\dTlsDot\dFromFrameWorldToIMU}
        \def\dTlsDotFromIMUToIndivLiDAR{\dTlsDot\dFromFrameIMUToIndivLiDAR}

    \def\dTlsTrans{\mathbf{t}^{\top}}
    \newcommand{\nTlsTransUO}[1]{\dTlsTrans_{#1}} %

        \def\dTlsPert{\boldsymbol{\rho}}
        \newcommand{\nTlsPertUO}[1]{\dTlsPert_{#1}} %

            \def\dTlsPertSkew{\dTlsPert^{\wedge}}
            \newcommand{\nTlsPertSkewUO}[1]{\dTlsPertSkew_{#1}} %
            \def\dTlsPertTrans{\dTlsPert^{\top}}
            \newcommand{\nTlsPertTransUO}[1]{\dTlsPertTrans_{#1}} %

        \def\dRotPert{\boldsymbol{\phi}}
        \newcommand{\nRotPertUO}[1]{\dRotPert_{#1}} %
            \def\dRotPertSkew{\dRotPert^{\wedge}}
            \newcommand{\nRotPertSkewUO}[1]{\dRotPertSkew_{#1}} %
            \def\dRotPertTrans{\dRotPert^{\top}}
            \newcommand{\nRotPertTransUO}[1]{\dRotPertTrans_{#1}} %

\def\dLthLiDARFrame{\dLiDARFrame^{l}}
\def\dLthIMUFrame{\dIMUFrame^{l}}

\def\dIthLiDARFrame{\dLthLiDARFrame_{i}}
\def\dJthLiDARFrame{\dLthLiDARFrame_{j}}
\def\dKthLiDARFrame{\dLthLiDARFrame_{k}}

\def\dIthOfLthLiDARFrame{\dLthLiDARFrame_{i}}
\def\dJthOfLthLiDARFrame{\dLthLiDARFrame_{j}}
\def\dKthOfLthLiDARFrame{\dLthLiDARFrame_{k}}

    \def\dLiDARFrame{\mathit{L}}
    \def\dIndivLiDARFrame{\dLiDARFrame^{l}}
        \def\dJthIndivLiDARFrame{\dIndivLiDARFrame_{j}}
        \def\dKthIndivLiDARFrame{\dIndivLiDARFrame_{k}}
    \def\dIMUFrame{\mathit{I}}
        \def\dIthIMUFrame{\dIMUFrame_{i}}
        \def\dKthIMUFrame{\dIMUFrame_{k}}
    \def\dIndivIMUFrame{\dIMUFrame^{l}}
        \def\dJthIndivIMUFrame{\dIndivIMUFrame_{j}}
        \def\dKthIndivIMUFrame{\dIndivIMUFrame_{k}}
    \def\dWorldFrame{\mathit{W}}

\def\dFromFrameWorldToIMU{_{\dWorldFrame \dIMUFrame}}

\def\dFromFrameIMUToIndivLiDAR{_{\dIMUFrame \dIndivLiDARFrame}}
\def\dFromFrameKthIMUToJthIndivIMU{_{\dKthIMUFrame \dJthIndivIMUFrame}}
\def\dFromFrameKthIMUToJthIndivLiDAR{_{\dKthIMUFrame \dJthIndivLiDARFrame}}

\def\dFromFrameWorldToKthIMU{_{\dWorldFrame \dKthIMUFrame}}
\def\dFromFrameWorldToIthIMU{_{\dWorldFrame \dIthIMUFrame}}
\def\dFromFrameWorldToJthIndivIMU{_{\dWorldFrame \dJthIndivIMUFrame}}
\def\dFromFrameWorldToJthIndivLiDAR{_{\dWorldFrame \dJthIndivLiDARFrame}}

\def\dJthIMUFrame{\dLthIMUFrame_{j}}

\def\dFromIdxAToB{_{a b}}
\def\dFromIdxBToC{_{b c}}
\def\dFromIdxAToC{_{a c}}

\def\dFromIdxIToJ{_{i j}}
\def\dFromIdxJToK{_{j k}}
\def\dFromIdxIToK{_{i k}}
\def\dFromIdxJToI{_{j i}}
\def\dFromIdxKToJ{_{k j}}
\def\dFromIdxKToI{_{k i}}

\def\dInJthIndivLiDARFrame{ {}^{\dJthIndivLiDARFrame}}

\def\dInKthIMUFrame{ {}^{\dKthIMUFrame}}
\def\dInJthIMUFrame{ {}^{\dJthIMUFrame}}
\def\dInJthIndivIMUFrame{ {}^{\dJthIndivIMUFrame}}

\def\dInWorldFrame{ {}^{\dWorldFrame}}

\def\dFromIjToLj{_{\dIthIMUFrame \dJthIMUFrame}}
\section{Introduction} \label{sec:introduction}
Simultaneous localization and mapping (SLAM)~\cite{song2022ral} using 3D LiDAR sensors has been actively applied in robotic applications owing to their accuracy and reliability~\cite{ignacio2023lra},~\cite{behley2018rss}.
Despite significant advancements in LiDAR-based SLAM methods~\cite{lim2023UR}, traditional single LiDAR systems are susceptible to perspective limitations, particularly due to limited field of view (FOV) and occlusion.

To address this issue, multiple LiDAR systems have recently attracted research interest~\cite{Jiao2022tro},~\cite{Reinke2022ral},~\cite{Jung2023ral}.
However, adopting multiple LiDAR systems can be challenging due to the discrepancies in measurement noise across different LiDARs.
Different LiDARs will have different measurement noises, which arises from the variations in manufacturers or scanning patterns~\cite{Yuan2021ral}.
Existing methods have typically overlooked the noise discrepancies, treating the measurement noise as homogeneous, which can significantly affect accuracy.

To tackle the challenge above, we propose Noise-aware Multiple LiDAR-Inertial Odometry (NM-LIO) that addresses the noise discrepancies. 
We integrate a noise model to quantify the measurement noise of each LiDAR.
Additionally, we estimate the uncertainty of the residuals based on the measurement noise during the state update process, allowing the measurement model to capture the noise discrepancies.
We evaluate the proposed method on a public multiple LiDAR dataset~\cite{Helmberger2022ral} and compare it with state-of-the-art methods.
The experimental results demonstrate that the proposed method can accurately estimate the odometry in various environments by accounting for the noise discrepancies.

The remainder of this paper is organized as follows:
In Section~\ref{sec:related_works}, we provide a review of related works.
In Section~\ref{sec:methodology}, we describe an overview of our proposed multiple LiDAR-inertial odometry system and the details of each key process. 
The experimental results are presented in Section~\ref{sec:experiments} followed by the conclusions in Section~\ref{sec:conclusion}.
\section{Related Works} \label{sec:related_works}
Fast-LIO~\cite{wei2021ral} and Fast-LIO2~\cite{wei2022tro} are tightly-coupled iterated error state Kalman filter (IESKF)-based LiDAR-inertial odometry (LIO) systems that serve as foundational frameworks for numerous LIO systems.
Fast-LIO enhances computational efficiency by introducing a novel Kalman gain formula that reduces the computation complexity from the dimension of the measurements to the dimension of the state. 
Additionally, Fast-LIO2 introduces an efficient incremental kd-tree data structure that enables dynamic insertion and rebalancing.
Several related works have incorporated advanced techniques such as adaptive downsampling~\cite{lim2023UR} and efficient local map management~\cite{chunge2022ral} to improve the accuracy and robustness of the odometry.

Due to the limited FOV and occlusion of single LiDAR, multiple LiDAR systems have recently attracted research interest.
M-LOAM~\cite{Jiao2022tro} proposes a sliding window-based multiple LiDAR odometry that simultaneously estimates poses and calibration parameters.
It computes the point-wise uncertainty by composing the state covariance and the extrinsic covariance.
MA-LIO~\cite{Jung2023ral} addresses the discrepancies in terms of temporal and spatial factors inherent in multiple LiDAR systems.
It mitigates these discrepancies by employing B-spline interpolation as an alternative for linear interpolation and accurately transferring points among LiDARs.
Additionally, it propagates point-wise uncertainty throughout the motion compensation process, assigning uncertainty caused by the motion to each point.
However, existing methods treat the measurement noise of each LiDAR as homogeneous, neglecting the discrepancies that can significantly affect the accuracy.

Several recent single LIO systems have addressed the measurement noise. 
VoxelMap~\cite{Yuan2022ral} incorporates the noise model from~\cite{Yuan2021ral}, which estimates the noise based on uncertainties in both bearing direction and range measurements, to improve the robustness of odometry. 
It composes measurement noise with pose uncertainty to represent point uncertainty, which is then propagated to the plane parameters, enabling a probabilistic representation of the environment. 
LOG-LIO2~\cite{huang2024ral} further extends this approach by introducing a novel point uncertainty model that considers both measurement noise~\cite{Yuan2021ral} and surface characteristics.
In our work, we utilize the measurement noise model from~\cite{Yuan2021ral} to quantify the measurement noise for each LiDAR. 
Additionally, by estimating the uncertainty of the residuals based on the measurement noise, we can capture the noise discrepancies across different LiDARs.
\section{Multiple LiDAR-Inertial Odometry With Measurement Noise Model}\label{sec:methodology}
\subsection{Overview} \label{subsec:overview}
\begin{figure}[!t]
    \centering
    \includegraphics[width=\textwidth]{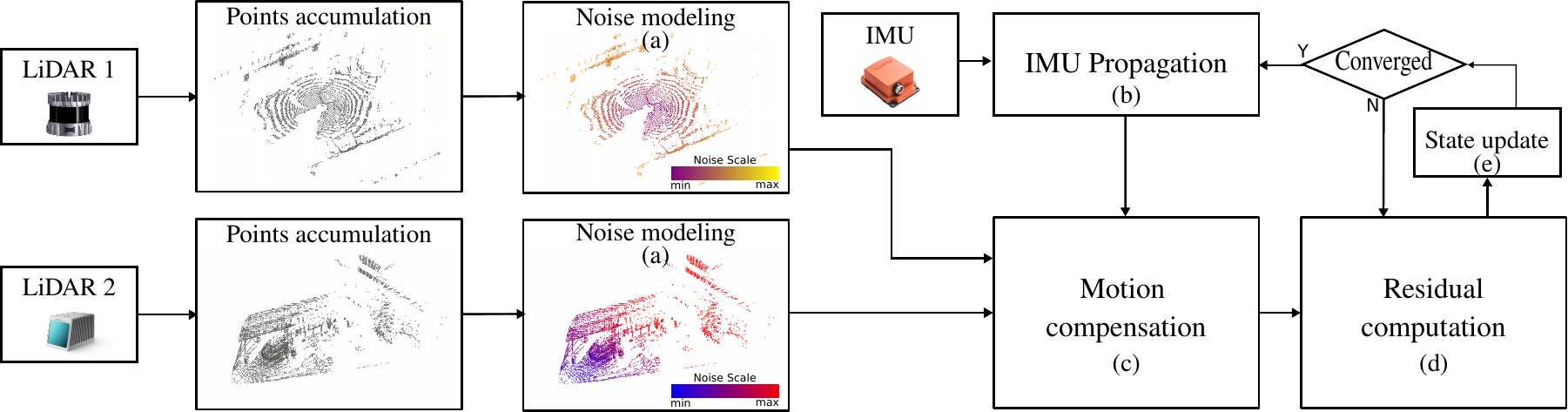}
    \caption{Overview of our proposed Noise-aware Multiple LiDAR-Inertial Odometry (NM-LIO) system.
    The system consists of five key processes. 
    (a) First, LiDAR measurement noise is modeled after the points are accumulated.
    (b) Second, IMU propagation is performed to predict the state while the LiDAR points are accumulated.
    (c) Third, motion compensation is performed to compensate for the motion distortion.
    (d) Fourth, the residual is computed to form the measurement model.
    (e) Finally, the state is iteratively updated until it converges.
    }
    \label{fig:framework}
\end{figure}

In this section, we will describe our proposed Noise-aware Multiple LiDAR-Inertial Odometry (NM-LIO) system, which addresses the noise discrepancies between LiDARs.
The overview of the proposed system is shown in Fig.~\ref{fig:framework}.
In the following subsections, we will first introduce the system's kinematic model and the random variables associated with the state, followed by a detailed description of each process.
\subsection{Kinematic Model} \label{subsec:kinematic_model}

Taking the IMU frame (denoted as $\dIMUFrame$) as the body frame of reference leads to following kinematic model:
\begin{equation}
    \begin{aligned}
        \dTlsDotFromWorldToIMU &= \nVelUO{\dWorldFrame \dIMUFrame} \;, \quad
        \dRotDotFromWorldToIMU = (\dGyroMeas-\dGyroBias-\dGyroNoise)^{\wedge}, \\
        \nVelDotUO{\dWorldFrame \dIMUFrame} &= \nRotUO{\dWorldFrame \dIMUFrame} (\dAccMeas-\dAccBias-\dAccNoise) + \dGrav, \\
        \dGyroBiasDot &= \dGyroBiasNoise \;, \quad
        \dAccBiasDot = \dAccBiasNoise \;, \quad
        \dGravDot = \mathbf{0},
    \end{aligned}
    \label{eq:continuous_model}
\end{equation}
where $\nRotUO{\dWorldFrame \dIMUFrame}$, $\nTlsUO{\dWorldFrame \dIMUFrame}$ are the rotation and translation of IMU in the world frame (denoted as $\dWorldFrame$), respectively. $\nVelUO{\dWorldFrame \dIMUFrame}$ is the velocity of IMU in the world frame,
$\dOmega_{m}$ and $\dAcc_{m}$ are the IMU inputs, $\nNoiseUO{\dOmega}$ and $\nNoiseUO{\dAcc}$ are IMU noise, $\nBiasUO{\dOmega}$ and $\nBiasUO{\dAcc}$ are the IMU biases,
$\nNoiseUO{\nBiasUO{\dOmega}}$ and $\nNoiseUO{\nBiasUO{\dAcc}}$ are the IMU bias noise, and $\dGrav $ is the gravity vector in the world frame.
$(\cdot)^{\wedge}$ denotes the skew-symmetric matrix of a vector.
Based on the $\boxplus$ operator in~\cite{He2023tie}, \eqref{eq:continuous_model} can be discretized as:
\begin{equation} %
    \mathbf{x}_{i+1} = \mathbf{x}_i \boxplus (\Delta t \mathbf{f}(\dIthStateEst, \mathbf{u}_i, \nProcNoiseUO{i})),
    \label{eq:discrete_model_form}
\end{equation}
where $\mathit{i}$ is the IMU index.
The function $\mathbf{f}$ is defined as:
\begin{equation} %
    \begin{aligned}
        &\mathbf{f}(\mathbf{x}, \mathbf{u}, \mathbf{w}) = \begin{bmatrix}
            \nVelUO{\dWorldFrame \dIMUFrame}  + \frac{1}{2}( \nRotUO{\dWorldFrame \dIMUFrame} (\dAcc-\nBiasUO{\dAcc}-\nNoiseUO{\dAcc}) + \dGrav ) \Delta t \\
            \dOmega_{m}-\nBiasUO{\dOmega}-\nNoiseUO{\dOmega} \\
            \nRotUO{\dWorldFrame \dIMUFrame} (\dAcc-\nBiasUO{\dAcc}-\nNoiseUO{\dAcc}) + \dGrav \\
            \nNoiseUO{\nBiasUO{\dOmega}} \\
            \nNoiseUO{\nBiasUO{\dAcc}} \\
            \mathbf{0}_{3 \times 1} \\
        \end{bmatrix}
    \end{aligned}
    \label{eq:discrete_model_details}
\end{equation}
where state $\mathbf{x}$, input $\mathbf{u}$, and noise $\mathbf{w}$ are defined as:
\begin{equation}
    \begin{aligned}
        \dManifold  &\triangleq  \dR3 \times \dSO3  \times \mathbb{R}^{\text{12}}  \\
        \mathbf{x} &\triangleq  [
            \nTlsTransUO{\dWorldFrame \dIMUFrame}, \; \nRotTransUO{\dWorldFrame \dIMUFrame} , \; \nVelTransUO{\dWorldFrame \dIMUFrame} , \; \nBiasTransUO{\dOmega} , \; \nBiasTransUO{\dAcc} , \; \dGravTrans 
        ]^{\top} \\
        \mathbf{u} &\triangleq [\dAccMeasTrans , \dGyroMeasTrans]^{\top}, \\
        \dProcNoise &\triangleq [\nNoiseTransUO{\dAcc} , \nNoiseTransUO{\dOmega} , \nNoiseTransUO{\nBiasUO{\dAcc}} , \nNoiseTransUO{\nBiasUO{\dOmega}}]^{\top}, \\
    \end{aligned}
    \label{eq:state_description}
\end{equation}

\subsection{Error State  Kalman Filter}
Our proposed method is based on the IESKF~\cite{wei2021ral}.
We define random variables for the true state $\dState$ according to a `large' noise-free state $\dStateBar$ and a `small' noisy perturbation $\dStateErr$ via $\boxplus$ operation as below:
\begin{equation} %
    \dState \triangleq \dStateBar\boxplus \dStateErr. \\
    \label{eq:state_decomposition}
\end{equation}
$\dStateErr$ is defined as
\begin{equation} %
    \dStateErr \triangleq  [ \dTPertTransFromWorldToIMU, \; \dTPertTransFromIMUToIndivLiDAR, \dVelErrTransFromWorldToIMU, \; \dGyroBiasErrTrans, \; \dAccBiasErrTrans, \; \dGravErrTrans ]^{\top}, \\
    \label{eq:true_state}
\end{equation}
where $\dTPert = \dBigLog(\dTBarInv \dT)$ is the transformation error and the others are additive errors $\mathbf{\widetilde{x}} = \mathbf{x} - \bar{\mathbf{x}}$.
By defining $\dStateErr$ to be a Gaussian random variable with zero mean in the Lie algebra, $\dStateErr \sim \mathcal{N}(\mathbf{0}, \dStateCov)$, we induce a probability distribution function over $\dManifold$ that is parameterized with a mean $\dStateBar \in \dManifold$ and a covariance $\dStateCov$ defined in the Lie algebra~\cite{Mangelson2020tro}.

\subsection{IMU Propagation} \label{subsec:imu_propagation}
IMU measurements are accumulated while the LiDAR point cloud is being processed.
The IMU propagation is performed following \eqref{eq:discrete_model_form} as:
\begin{equation} %
    \dStateEst_{i+1} = \dIthStateEst \boxplus (\Delta t \mathbf{f}(\dIthStateEst, \mathbf{u}_i, \dZero)) \;\; ; \;\; \dStateEst_{0} = \dStateEst_{k-1},
    \label{eq:state_prediction}
\end{equation}
where $\Delta t$ is the time interval between the IMU measurements.
Propagated error state is derived as:
\begin{equation} %
    \begin{aligned}
        \nStateErrUO{i+1} &= \nStateUO{i+1} \boxminus \dStateEst_{i+1} \\
        &= (\nStateUO{i} \boxplus \Delta t \mathbf{f}(\dIthStateEst, \mathbf{u}_i, \nProcNoiseUO{i})) \boxminus \dStateEst_{i+1} \\
        &= ((\dIthStateEst \boxplus \nStateErrUO{i}) \boxplus \Delta t \mathbf{f}(\dIthStateEst, \mathbf{u}_i, \nProcNoiseUO{i})) \boxminus \dStateEst_{i+1}. \\
    \end{aligned}
    \label{eq:error_state_derivation}
\end{equation}
If we let the right-hand side of \eqref{eq:error_state_derivation} to be $\dIMUPropJaco(\nStateErrUO{i},\nProcNoiseUO{i})$, then the error state could be approximated as
\begin{equation} %
    \begin{aligned}
        \nStateErrUO{i+1} &\approx \dIMUPropJaco(\dZero,\dZero) + \nIMUPropJacoUO{\nStateErrUO{i}}\nStateErrUO{i} + \nIMUPropJacoUO{\nProcNoiseUO{i}} \nProcNoiseUO{i},\\
    \end{aligned}
    \label{eq:error_state_approx}
\end{equation}
where each Jacobian is defined as
\begin{equation}
    \begin{aligned}
        \nIMUPropJacoUO{\nStateErrUO{i}} = \frac{\partial \dIMUPropJaco(\nStateErrUO{i},\nProcNoiseUO{i})}{\partial \nStateErrUO{i}} \Bigg|_{\nStateErrUO{i} = 0, \nProcNoiseUO{i} = 0} \;, \quad
        \nIMUPropJacoUO{\nProcNoiseUO{i}} = \frac{\partial \dIMUPropJaco(\nStateErrUO{i},\nProcNoiseUO{i})}{\partial \nProcNoiseUO{i}} \Bigg|_{\nStateErrUO{i} = 0, \nProcNoiseUO{i} = 0}.
    \end{aligned}
\end{equation}
Finally, the propagated error state, $\nStateErrUO{i+1}$, and corresponding covariance matrix, $\nStateCovEstUO{i+1}$, are obtained as following equations:
\begin{equation}
    \begin{aligned}
        \nStateErrUO{i+1} &= \nIMUPropJacoUO{\nStateErrUO{i}} \nStateErrUO{i} + \nIMUPropJacoUO{\nProcNoiseUO{i}} \nProcNoiseUO{i} , \;\; \nStateErrUO{0} = \nStateErrUO{k-1} \\
        \nStateCovEstUO{i+1} &= \nIMUPropJacoUO{\nStateErrUO{i}} \nStateCovEstUO{i} \nIMUPropJacoTransUO{\nStateErrUO{i}} + \nIMUPropJacoUO{\nProcNoiseUO{i}} \nProcNoiseCovUO{i} \nIMUPropJacoTransUO{\nProcNoiseUO{i}} , \;\; \nStateCovEstUO{0} = \nStateCovBarUO{k-1}.
    \end{aligned}
    \label{eq:errorstate_imupropagation}
\end{equation}
\subsection{Motion Compensation}
LiDAR typically emits pulsed light waves into the environment sequentially.
Each LiDAR points is sampled at a point rate (e.g. 200kHz) and accumulated into a single scan at a scan rate (e.g. 10Hz).
The sensor undergoes continuous motion while the scan is being accumulated, creating significant motion distortion.
To compensate for this motion distortion, IMU measurements can be used.
The state at the point sampling time can be obtained as
\begin{equation} %
    \begin{aligned}
        \nStateUO{j} &= \nStateUO{i} \boxplus ( \Delta t \dIMUPropJaco(\dIthStateEst, \mathbf{u}_i, \nProcNoiseUO{i})), \quad \Delta t = \rho_j - \tau_i, \\
    \end{aligned}
    \label{eq:state_in_LiDAR_frame}
\end{equation}
where $\rho_j$ is the point sampling time of the $j$-th LiDAR bounded $[\tau_i, \tau_{i+1}]$.
By transforming to the local pose at the last point time, the motion distortion is compensated using
\begin{equation} %
    \dJthPtHomoInKthIMUFrame = \nTUO{\dKthIMUFrame \dJthIMUFrame} \dTFromIMUToIndivLiDAR \dJthPtHomoInJthIndivLiDARFrame
    \label{eq:LiDAR_compensated_point}
\end{equation}
where $\dPtHomo$ denotes the homogeneous coordinates of the LiDAR point to make compatible with the transformation matrix (i.e., $\dPtPrimeHomo = \dT \dPtHomo$).
Finally, the compensated LiDAR points form a scan whose points can be assumed as all sampled at time $t_k$.
\def\dBearingDirInLthLiDARFrame{{\dBearingDir}^{l}}
\def\dBearingDirEstInLthLiDARFrame{{\dBearingDirEst}^{l}}
\def\dBearingDirErrInLthLiDARFrame{{\dBearingDirErr}^{l}}
\def\dRangeInLthLiDARFrame{{\dRange}^{l}}
\def\dRangeEstInLthLiDARFrame{{\dRangeEst}^{l}}
\def\dRangeErrInLthLiDARFrame{{\dRangeErr}^{l}}

\begin{figure}[t!]
    \centering
    \begin{subfigure}[b]{0.45\textwidth}
        \centering
        \includegraphics[width=\textwidth]{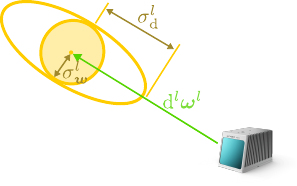}
        \caption{}
        \label{fig:noise_model_a}
    \end{subfigure}
    \hspace{0.1cm}
    \begin{subfigure}[b]{0.45\textwidth}
        \centering
        \includegraphics[width=\textwidth]{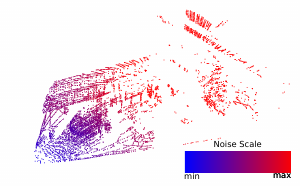}
        \caption{}
    \end{subfigure}
    \caption{
        Illustration of the LiDAR measurement noise model. 
        (a) Measurement noise is modeled based on uncertainties in range and bearing direction.
        (b) Measurement noise is visualized by a color scale.
    }
    \label{fig:noise_model}
\end{figure}
\subsection{Measurement Noise Model for Multiple LiDARs} \label{subsec:point_uncertainty_modeling_propagation}

We quantify the measurement noise of each LiDAR using a noise model from~\cite{Yuan2021ral}.
The uncertainty of a point is modeled based on the uncertainties in range and bearing direction.
Let $\dBearingDirInLthLiDARFrame$ be the bearing direction of the $l$-th LiDAR point and $\dBearingDirErrInLthLiDARFrame$ be the its noise,
the true bearing direction $\dBearingDirInLthLiDARFrame$ is obtained using the $\boxplus$ operation encapsulated in $\dS2$ as below:
\begin{equation}
    \dBearingDirInLthLiDARFrame = \dBearingDirEstInLthLiDARFrame
    \nBoxPlusUO{\dS2} \dBearingDirErrInLthLiDARFrame
    \triangleq e^{(\mathbf{N}(\dBearingDirEstInLthLiDARFrame) \dBearingDirErrInLthLiDARFrame )^{\wedge}} \dBearingDirEstInLthLiDARFrame,
    \label{eq:bearing_direction}
\end{equation}
where $\mathbf{N}(\dBearingDirEst) = [\mathbf{N}_1, \mathbf{N}_2] \in \mathbb{R}^{3 \times 2}$ is an orthonormal basis of the tangent plane at $\dBearingDirEst$.
Similarly, let $\dRangeInLthLiDARFrame$ be the depth measurement and $\dRangeErrInLthLiDARFrame$ be its noise,
then the relation between the true depth $\dRangeInLthLiDARFrame$ and its measurement $\dRangeEstInLthLiDARFrame$ is
\begin{equation}
    \dRangeInLthLiDARFrame = \dRangeEstInLthLiDARFrame + \dRangeErrInLthLiDARFrame.
    \label{eq:range}
\end{equation}
Here, $\dBearingDirErrInLthLiDARFrame$ and $\dRangeErrInLthLiDARFrame$ are assumed to be Gaussian noise as:
\begin{equation}
    \begin{aligned}
        \dBearingDirErrInLthLiDARFrame &\sim \nNormalDist{\dZero}{\dCov_{\dBearingDirInLthLiDARFrame}} \;, 
        \quad \dCov_{\dBearingDirInLthLiDARFrame} = 
        \begin{bmatrix}
            \sigma_{\dBearingDirInLthLiDARFrame}^{2} & 0 \\
            0 & \sigma_{\dBearingDirInLthLiDARFrame}^{2}
        \end{bmatrix}, \\
        \dRangeErrInLthLiDARFrame &\sim \nNormalDist{\dZero}{\dCov_{\dRangeInLthLiDARFrame}} \;,
        \quad \dCov_{\dRangeInLthLiDARFrame} = \sigma_{\dRangeInLthLiDARFrame}^2,
    \end{aligned}
    \label{eq:noise_distribution}
\end{equation}
where $\sigma_{\dBearingDirInLthLiDARFrame}$ and $\sigma_{\dRange}$ are the standard deviations of the bearing direction and range noise, respectively.
For the point of $l$-th LiDAR $\dPt^{l}$, let $\dNoise^{l}$ be its measurment noise.
Combining \eqref{eq:bearing_direction} and \eqref{eq:range}, we can obtain the relation between the true point $\dPt^{l}$ and its measurement $\dPtEst^{l}$ as
\begin{equation}
    \begin{aligned}
    \dPt^{l} &= \dRangeInLthLiDARFrame \dBearingDirInLthLiDARFrame \\
    \dPtEst^{l} + \dNoise^{l}  &= (\dRangeEstInLthLiDARFrame + \dRangeErrInLthLiDARFrame) (\dBearingDirEstInLthLiDARFrame \boxplus_{\dS2} \dBearingDirErrInLthLiDARFrame) \\
    &\approx \dRangeEstInLthLiDARFrame \dBearingDirEstInLthLiDARFrame + \dBearingDirEstInLthLiDARFrame \dRangeErrInLthLiDARFrame - \dRangeEstInLthLiDARFrame \dBearingDirEst^{l \wedge} \mathbf{N}(\dBearingDirEstInLthLiDARFrame) \dBearingDirErrInLthLiDARFrame.
    \end{aligned}
    \label{eq:point_uncertainty_in_lidar_frame}
\end{equation}
Then, we can model the point uncertainty in the local LiDAR pose as below:
\begin{gather}
    \dNoise^{l} = \underbrace{[\dBearingDirEstInLthLiDARFrame \quad - \dRangeEstInLthLiDARFrame \dBearingDirEst^{l \wedge} \mathbf{N}(\dBearingDirEst)]}_{\mathbf{A}} \begin{bmatrix}
        \dRangeErrInLthLiDARFrame \\
        \dBearingDirErrInLthLiDARFrame 
    \end{bmatrix} \sim \nNormalDist{\dZero}{\dCov_{\dPt^{l}}}, \\
    \dCov_{\dPt^{l}} = \mathbf{A} \begin{bmatrix}
        \dCov_{\dRange^{l}} & \dZero_{1 \times 2} \\
        \dZero_{2 \times 1} & \dCov_{\dBearingDir^{l}}
    \end{bmatrix} \mathbf{A}^{\top},
    \label{eq:point_uncertainty_covariance}
\end{gather}
where $\dCov_{\dRange^{l}}$ and $\dCov_{\dBearingDir^{l}}$ are the uncertainties of the range and bearing direction, respectively.
Illustration of the measurement noise model is shown in Fig.~\ref{fig:noise_model}. 
Fig.~\ref{fig:noise_model_a} is adapted and modified from Fig. 1a from~\cite{huang2024ral}.
By properly setting the standard deviation of range and bearing direction for each LiDAR, which are $\sigma_{\dRangeInLthLiDARFrame}$ and $\sigma_{\dBearingDirInLthLiDARFrame}$, respectively, 
the merged frame can capture the measurement noise discrepancies between LiDARs.
\subsection{Measurement Model}
We further transform the LiDAR scan obtained through motion compensation into the world frame and form residuals.
The measurement noise from section~\ref{subsec:point_uncertainty_modeling_propagation} should also be propagated into the world frame by composing with the uncertainty of the predicted pose from section \ref{subsec:imu_propagation}. 
We define the residual based on the assumption that the true point lies on a near plane in the local map as below:
\begin{equation} %
    \dZero = (\mathbf{D} \dJthPlaneNormal)^{\top}(\dTFromFrameWorldToKthIMU \dJthPtHomoInKthIMUFrame - \mathbf{D} \dJthPlaneCenter)
    \label{eq:measurement_model}
\end{equation}
where $ \dJthPlaneNormal$ and $\dJthPlaneCenter$ are the normal vector and center of the local plane, respectively, and $\mathbf{D}$ is the dilation matrix.
We define the compact form of the measurement model and its first-order approximation as below:
\begin{equation}
    \begin{aligned}
        \dZero &= \mathbf{h}_{j} (\dKthStateEstIter \boxplus \dKthStateErrIter , \dJthNoiseInKthIMUFrame) \\
        & \simeq \mathbf{h}_{j} ( \dKthStateEstIter, \dZero ) + \mathbf{H}^{\kappa}_{j} \dKthStateErrIter + \dJthMeasNoise.
    \end{aligned}
    \label{eq:measurement_model_approx}
\end{equation}
$\mathbf{H}^{\kappa}_{j}$ is the Jacobian matrix of measurement model with respect to $\dKthStateErrIter$, evaluated at zero, and $\dJthMeasNoise$ is the uncertainty of the residual as below:
\begin{gather}
    \mathbf{H}^{\kappa}_{j} = \begin{bmatrix}
        ((\mathbf{D} \dJthPlaneNormalIter)^{\top} \dTEstIterFromWorldToIMU \dJthPtEstHomoInKthIMUFrame)^{\top} \\
        \dZero_{1 \times 12}^{\top}
    \end{bmatrix}^{\top} \\
    \dJthMeasNoise = \mathbf{J}_{\dJthNoiseInKthIMUFrame} \dJthNoiseInKthIMUFrame \; \quad \mathbf{J}_{\dJthNoiseInKthIMUFrame} = (\mathbf{D} \dJthPlaneNormalIter)^{\top} \dTEstIterFromWorldToIMU.
    \label{eq:measurement_jacobian}
\end{gather}
Note that the uncertainty of the residual $\dJthMeasNoise$ comes from the raw measurement noise $\dJthNoiseInKthIMUFrame$, enabling the measurement model to capture the noise discrepancies between LiDARs.
Finally, state update is performed by minimizing the sum of the prior distribution from \eqref{eq:errorstate_imupropagation} and the measurement model from \eqref{eq:measurement_model_approx}~\cite{wei2021ral}.

\section{Experimental Results} \label{sec:experiments} 
\begin{figure}[h!]
    \centering
    \begin{subfigure}[b]{0.45\textwidth}
        \centering
        \includegraphics[width=\textwidth]{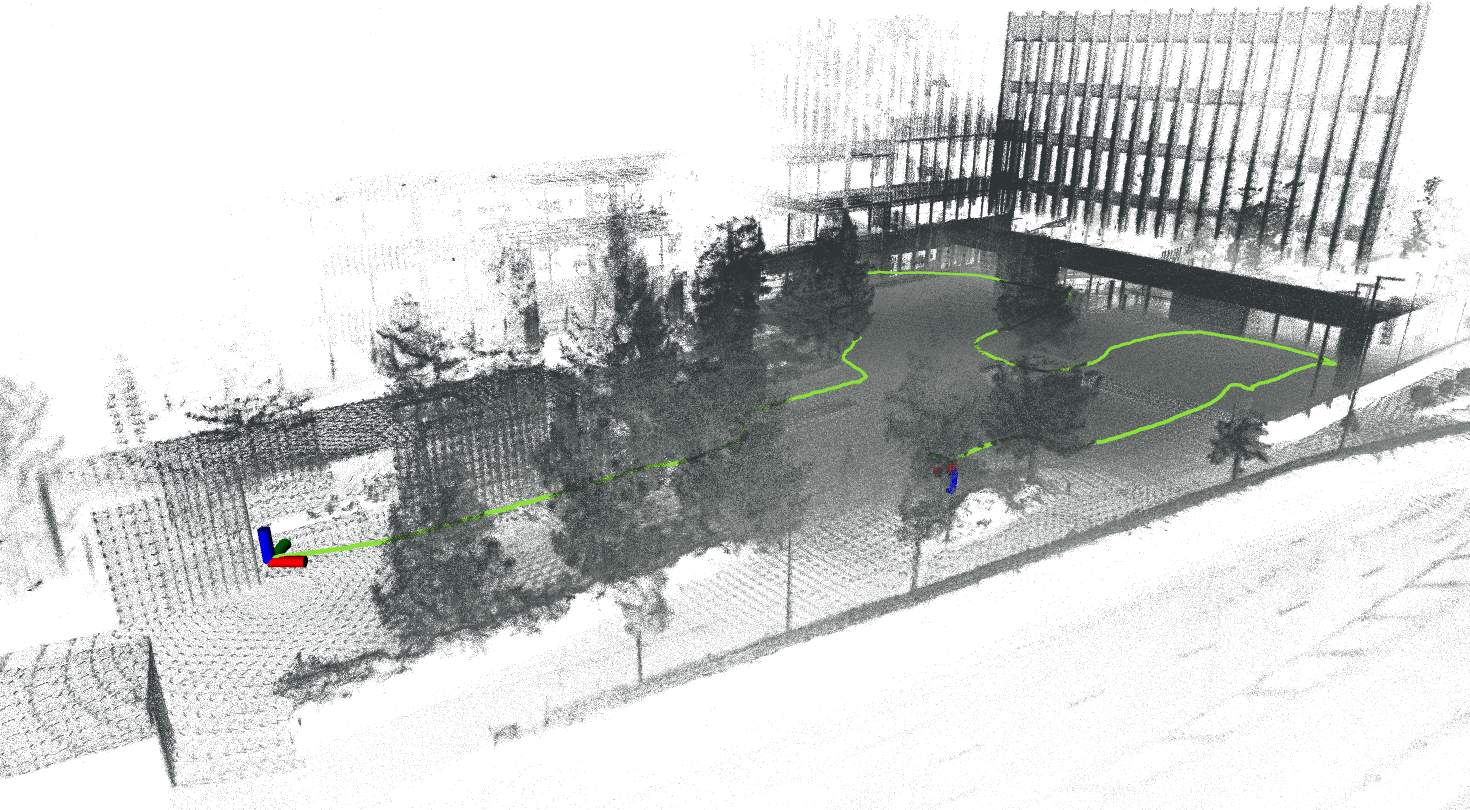}
        \caption{}
    \end{subfigure}
    \hspace{0.1cm}
    \begin{subfigure}[b]{0.45\textwidth}
        \centering
        \includegraphics[width=\textwidth]{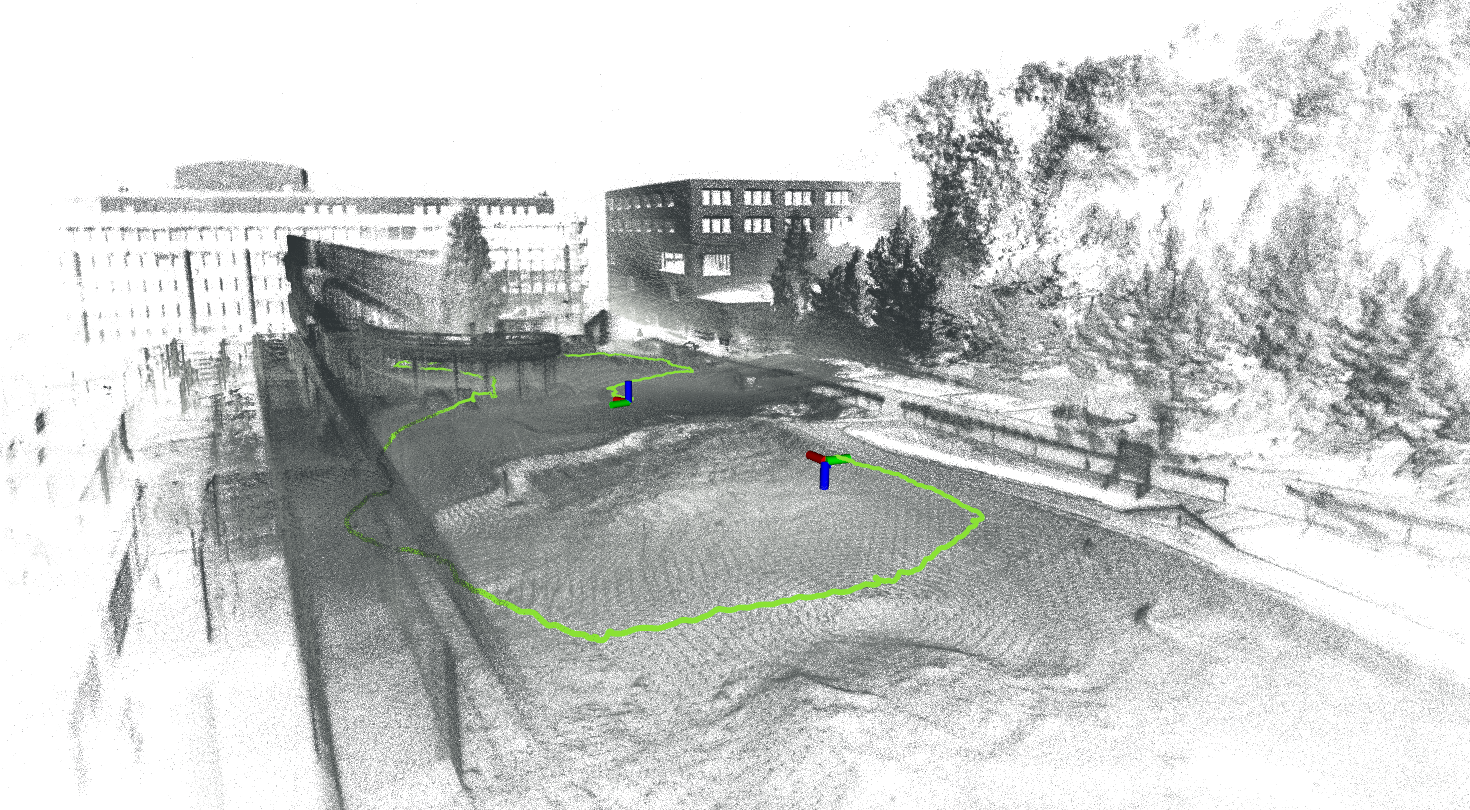}
        \caption{}
    \end{subfigure}
    \caption{
        Accumulated scan with estimated trajectory of the proposed method on the (a) campus sequence and the (b) construction sequence of the Hilti dataset~\cite{Helmberger2022ral}.
    }
    \label{fig:Campus_construction}
\end{figure}

Proposed method is evaluated on the Hilti SLAM dataset 2021~\cite{Helmberger2022ral}, which features a multiple LiDAR system consisting of an Ouster OS0-64 LiDAR and a Livox MID70 LiDAR.
The dataset provides various sequences of indoor and outdoor environments.
Proposed method is compared against three state-of-the-art multiple LiDAR odometry methods: M-LOAM~\cite{Jiao2022tro}, LOCUS2.0~\cite{Reinke2022ral}, and MA-LIO~\cite{Jung2023ral}.
Since the dataset provides the translation component of the ground truth, we only calculate the translation component of the absolute trajectory error (ATE).
The evaluation results are shown in Table~\ref{tab:hilti_exp}.
The results show that the our NM-LIO outperforms other state-of-the-art methods in various environments by incorporating the noise discrepancies between LiDARs. 
The accumulated scan with the estimated trajectory of the proposed method on the campus and construction sequences is shown in Fig.~\ref{fig:Campus_construction}.

\begin{table}[!h]
	\centering
    \caption{Translation component of ATE on the Hilti dataset~\cite{Helmberger2022ral}.}
	\setlength{\tabcolsep}{10pt}
    {\scriptsize
    \begin{tabular}{c|ccccc}
        \toprule
        & \multicolumn{5}{c}{Sequence} \\
        \cmidrule(lr){2-6}
        Method                          & UZH               & Basement          & Campus            & Construction Site & LAB \\
        \cmidrule(lr){1-1} \cmidrule(lr){2-6} 
        M-LOAM~\cite{Jiao2022tro}       & 0.276             & 0.115             & 0.386             & 2.647             & 0.064 \\
        LOCUS 2.0~\cite{Reinke2022ral}  & 0.177             & 0.120             & 0.087             & 0.290             & 0.040 \\
        MA-LIO~\cite{Jung2023ral}       & 0.177             & 0.036             & 0.046             & 0.063             & \textbf{0.024} \\
        Ours                            & \textbf{0.174}    & \textbf{0.023}    & \textbf{0.038}    & \textbf{0.061}    & 0.025 \\
        \bottomrule
    \end{tabular}
    }
    \label{tab:hilti_exp}
\end{table}

\section{Conclusions} \label{sec:conclusion}
In this paper, we proposed Noise-aware Multiple LiDAR-Inertial Odometry (NM-LIO) that addresses the discrepancies in measurement noise across different LiDARs.
Experimental results demonstrated that our NM-LIO could accurately estimate the odometry in various environments by accounting for the noise discrepancies.
However, selecting the proper noise model parameters for each LiDAR requires empirical determination, which involves considerable manual effort.
Therefore, in future work, we plan to incorporate these parameters into the state vector and estimate them along with the state.
Additionally, we aim to enhance the accuracy of the odometry by incorporating advanced techniques such as ground segmentation~\cite{oh2022travel} and refining the matching technique~\cite{lim2022icra}.

\bibliographystyle{splncs}
\bibliography{URL-bib}

\end{document}